\documentclass{article}
\usepackage{spconf,amsmath,graphicx}

\usepackage{makecell}
\usepackage{color, colortbl}
\usepackage{xcolor}
\usepackage{booktabs}
\usepackage{amssymb}
\usepackage{multirow}
\usepackage{hyperref}
\usepackage[T1]{fontenc}

\title{PFDM: Parser-Free Virtual Try-on via Diffusion Model}
\name{Yunfang Niu\(^1\), Dong Yi\(^{1,2}\), Lingxiang Wu\(^{1,2,*}\), Zhiwei Liu\(^{1,2}\), Pengxiang Cai\(^{1,3}\), Jinqiao Wang\(^{1,2,3}\)\thanks{\(*\)Corresponding author. This work was supported by the National Key R\&D Program of China under Grant (No.  2022ZD0160601), National Natural Science Foundation of China under Grants (62306315), China Postdoctoral Science Foundation (2022M713363).}}

\address{\(^1\)Foundation Model Research Center, Institute of Automation, \\Chinese Academy of Sciences, Beijing, China\\
\(^2\)Wuhan AI Research, Wuhan, China\\
\(^3\)School of Artificial Intelligence, University of Chinese Academy of Sciences, Beijing, China}
\definecolor{LightCyan}{rgb}{0.88,0.95,1}
\begin{document}

\ninept

\maketitle

\begin{abstract}
Virtual try-on can significantly improve the garment shopping experiences in both online and in-store scenarios, attracting broad interest in computer vision. However, to achieve high-fidelity try-on performance, most state-of-the-art methods still rely on accurate segmentation masks, which are often produced by near-perfect parsers or manual labeling. To overcome the bottleneck, we propose a parser-free virtual try-on method based on the diffusion model (PFDM). Given two images, PFDM can ``wear'' garments on the target person seamlessly by implicitly warping without any other information. To learn the model effectively, we synthesize many pseudo-images and construct sample pairs by wearing various garments on persons. Supervised by the large-scale expanded dataset, we fuse the person and garment features using a proposed Garment Fusion Attention (GFA) mechanism. Experiments demonstrate that our proposed PFDM can successfully handle complex cases, synthesize high-fidelity images, and outperform both state-of-the-art parser-free and parser-based models. 

\end{abstract}

\begin{keywords}
Virtual try-on, diffusion models, implicit warping, high-resolution image synthesis.
\end{keywords}

\section{Introduction}
\label{sec:intro}
Given two images of a person and a garment, virtual try-on is to ``wear'' on the garment while keeping the person's pose and identity unchanged. It has attracted wide research attention with the potential to enhance the shopping experiences in e-commerce and metaverse~\cite{pallathadka2023applications, nagy2021consumer}. For example, virtual try-on allows users to quickly browse the fitting effects and can reduce the refund possibilities. 

Existing virtual try-on methods can be categorized into parser-based and parser-free. All diffusion-based methods~\cite{morelli2023ladi, zhu2023tryondiffusion, gou2023taming, baldrati2023multimodal} and most of Generative Adversarial Networks (GAN) based methods~\cite{xie2023gp, lee2022high, han2018viton, choi2021viton, morelli2022dress, hu2022spg, wang2018toward} belong to the former, the performance of which heavily rely on the parsing data, such as keypoints, segmentation maps, and etc. When noisy parsing results are met in practical scenarios, these methods are easy to fail. On the contrary, parser-free methods, such as~\cite{ge2021parser, lin2022rmgn, he2022style}, don't have this drawback. However, these GAN-based parser-free methods process the garment warping and blending in two separate steps, the training process of which are difficult and unstable. In addition, GAN-based methods are hard to synthesize high-resolution images and are prone to produce results with artifacts. In summary, there is no virtual try-on method that can generate high-resolution images in one step without parsing information.

To address the above issues, we introduce a \textbf{P}arser-\textbf{F}ree virtual try-on method based on \textbf{D}iffusion \textbf{M}odel \textbf{(PFDM)}. A comparison with the SOTA methods can be seen in Tab.~\ref{tab:1}. The proposed method has all advantages including high resolution, parser-free, and one-step pipeline. Specifically, we employ a denoising U-Net~\cite{ronneberger2015u} diffusion model to warp the garment to the target person in the latent space implicitly and restore the try-on images with an autoencoder~\cite{esser2021taming}. In the U-Net, we propose a try-on-specific Garment Fusion Attention (GFA) module, which can fuse the garment and person features in a multiscale and multihead way effectively. To further improve the robustness and generalization, we synthesize a large-scale training set as pseudo-inputs by wearing various garments on persons based on many existing models.

\begin{table}[t]
    \centering
    \caption{PFDM compare with the state-of-the-art virtual try-on methods. High resolution means that the resolution can reach 1024X768 (the resolution of images in VITON-HD~\cite{choi2021viton}).}
    \vspace{.12cm}
    \setlength{\tabcolsep}{.3em}
    \resizebox{\linewidth}{!}{
    \begin{tabular}{lc cccc}
    \toprule
    \textbf{Model} & & \textbf{Framework} & \makecell{\textbf{High}\\\textbf{Resolution}} & \makecell{\textbf{No}\\\textbf{Parser/Key-points}} & 
    \makecell{\textbf{Warping \& Rending}\\\textbf{Simultaneously}} \\

    \midrule
    HR-VTON~\cite{lee2022high} & & GAN & \checkmark &  & \\
    RMGN~\cite{lin2022rmgn} &  & GAN &  & \checkmark & \\
    Ladi-VTON~\cite{morelli2023ladi} &  & Diffusion &  &  & \\
    DCI-VTON~\cite{gou2023taming} &  & Diffusion & \checkmark &  & \\
    TryOnDiffusion~\cite{zhu2023tryondiffusion} & & Diffusion & \checkmark &  & \checkmark \\
    
    \rowcolor{LightCyan}
    \midrule
    \textbf{PFDM(ours)} & & Diffusion & \checkmark & \checkmark & \checkmark \\
    
    \bottomrule
    \end{tabular}
    }
    
\vspace{-.65cm}
\label{tab:1}
\end{table}

The key contributions of this paper can be summarized as follows: (1) Firstly, we propose a parser-free virtual try-on framework based on diffusion models. As we know, this is the first work to use diffusion models for parser-free virtual try-on. (2) Secondly, an enhanced cross-attention module is carefully designed to integrate person and garment features for implicit warping. (3) Finally, we evaluate our method on the VITON-HD~\cite{choi2021viton}, and the experiments show that our parser-free model can outperform the competitors by a consistent margin in both qualitative and quantitative evaluations.

\begin{figure*}[tp]
    \centering
    \includegraphics[width=\linewidth]{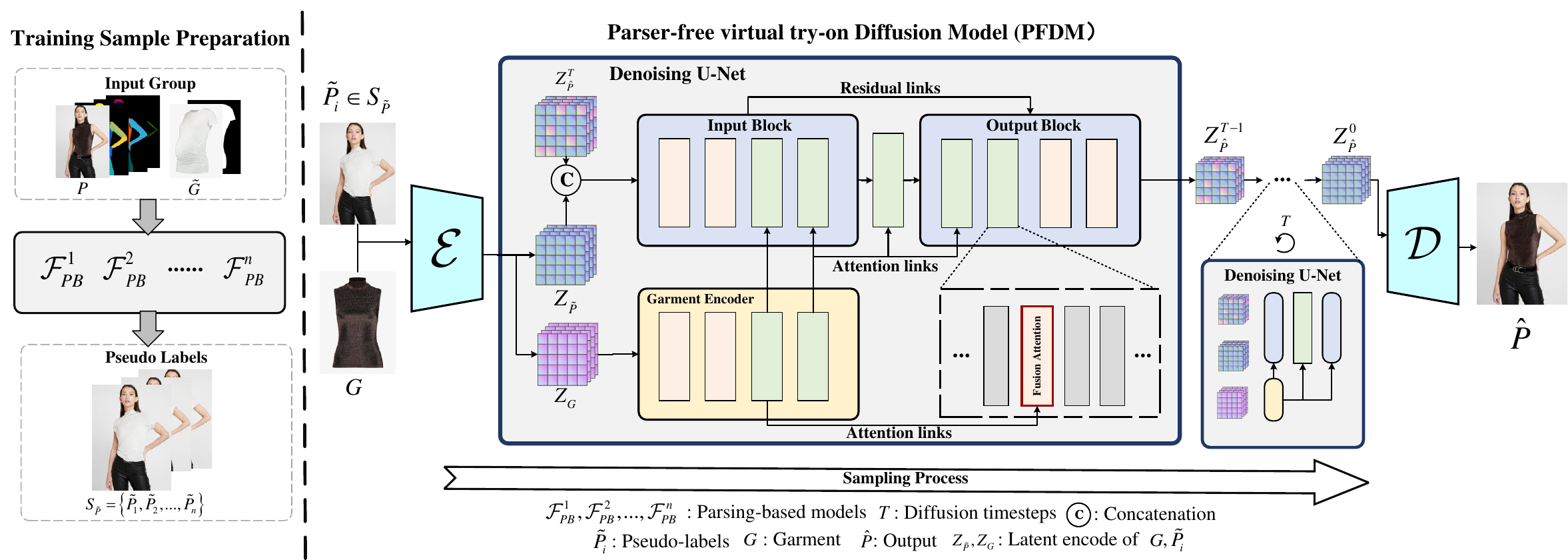}
    \caption{The architecture of proposed parser-free try-on method which contains two pipelines. First, we generate pseudo images \(\widetilde{P}\) of the person wearing another unpaired garment based on a model hub. Then, a well-designed diffusion model is trained to generate the try-on result \(\hat{P}\).}
    \label{fig:architecture}
\vspace{-.25cm}
\end{figure*}

\section{Related Work}
\label{sec:related}
\noindent\textbf{GAN for Virtual Try-on.} GAN-based virtual try-on methods usually adopt two-step architectures~\cite{xie2023gp, lee2022high, han2018viton, choi2021viton, wang2018toward, dong2022dressing, yang2022full, yu2019vtnfp} that first warp the garment to target shape, and then synthesize the results by combining the warped garment and the person images. Some works focus on enhancing the warping module based on thin-plate spline transformation (TPS) ~\cite{ge2021disentangled, morelli2022dress} or global flow~\cite{lee2022high, xie2023gp}. Other works aim to improve the performance of the generation module, e.g., adopting an alignment-aware generator to boost the resolution of the synthesized images~\cite{choi2021viton}, or refining the loss function to preserve the person's identity, body shape and pose~\cite{lewis2021tryongan}.

Most of the above GAN-based methods are generally parser-based. To dispense the hassle of using parsers for inference, WUTON ~\cite{issenhuth2020not} firstly proposed a parser-free try-on method using a student-teacher paradigm. Then several following works made improvements by knowledge distillation~\cite{ge2021parser}, introducing StyleGAN framework~\cite{he2022style} or designing regional mask feature fusion module~\cite{lin2022rmgn}. However, these models can only generate low-resolution ($< 512\times384$) images and are prone to produce results with artifacts.

\noindent\textbf{Diffusion Models for Virtual Try-on.}
Due to better training stability and mode coverage, diffusion models are more competitive than GANs~\cite{dhariwal2021diffusion, nichol2021improved} in image generation tasks. Therefore some cutting-edge works~\cite{morelli2023ladi, zhu2023tryondiffusion, baldrati2023multimodal} adopted diffusion models to blend the warped garment to the target person and achieve state-of-the-art performance. Ladi-VTON~\cite{morelli2023ladi} followed latent diffusion models and employed frozen VAE encoder-decoder to perform the diffusion process on latent space. However, it needed to prepare warped garments with additional warping modules and train the skip connection module to restore high-frequency details. TryOnDiffusion~\cite{zhu2023tryondiffusion} proposed a cross-attention mechanism for implicit warping between the streams of person and garment instead of channel-wise concatenation. Because the diffusion process in TryOnDiffusion is performed in pixel space, they need to train two try-on models and one super-resolution model to achieve satisfactory results. Besides that, these models both require pose or parsing information for inference.

\section{Proposed Method}
\label{sec:method}
 We propose a parser-free virtual try-on method that does not rely on the human parser and pose estimator for inference. Given a reference person \(\widetilde{P}\in\mathbb{R}^{3 \times H \times W}\) and a garment image \(G\in\mathbb{R}^{3 \times H \times W}\), our goal is to synthesis a try-on image \(\hat{P} \in \mathbb{R}^{3 \times H \times W}\) where the garment image \(G\) fits to the reference person \(\widetilde{P}\) and the non-garment regions in \(\widetilde{P}\) should be maintained. When training, \(\widetilde{P}\), \(G\) are model inputs and \(P\) is the synthesis target. Similar to other parser-free works~\cite{issenhuth2020not, ge2021parser, lin2022rmgn, he2022style}, \(\widetilde{P}\) is a pseudo-input obtained from \(P\) and unpaired garment \(\widetilde{G}\) by parser-based models, see Sec.~\ref{sec:3.1}. In particular, we design a garment encoder and a cross-attention module for feature fusion and implicit warping, see Sec.~\ref{sec:3.2}. The architecture of the proposed method is shown in Fig.~\ref{fig:architecture}.

\subsection{Parser-Free Virtual Try-on Diffusion Model}
\label{sec:3.1}
\noindent\textbf{Pseudo-input Preparation.} We need to prepare \({\widetilde{P}}\) for parser-free model training. Existing methods ~\cite{issenhuth2020not, ge2021parser, he2022style, lin2022rmgn} usually use only one model to get \(\widetilde{P}\). Differently, we choose a parser-based model hub (i.e., multiple models) \({\mathcal{F}^1_{PB}}\), \({\mathcal{F}^2_{PB}}\), ..., \({\mathcal{F}^n_{PB}}\) to obtain an image set \({\widetilde{P}_1}\), \({\widetilde{P}_2}\), ..., \({\widetilde{P}_n}\), which increases the diversity of the input data. The synthesis process is expressed as below
\begin{align}
{S_{\widetilde{P}}}=\{\widetilde{P}_i\}_{i=0}^n=\{\mathcal{F}^i_{PB}(P, I)\}_{i=0}^n
\end{align}
, where \(P\) is the target person for the parser-free model and \(I\) denotes the other inputs of the model hub including unpaired garment \(\widetilde{G}\), parsing mask \(\widetilde{M}\), skeleton \(\widetilde{K}\), and etc. 

Since the poses of the pseudo-inputs \(\widetilde{P}\) are well maintained, the subsequent try-on model can still learn to achieve good fitting effects when using the real-life person images \(P\) as the synthesis target. In addition, by introducing some noise carried by the pseudo-inputs, the robustness of the subsequent try-on model can be enhanced and lead the try-on results to exceed parser-based models.

\noindent\textbf{Diffusion Model.} The diffusion model could be represented as a Markov process where gradually adding noise to the data and converting it into a Gaussian noise. Then, reversibly restore the original data using a denoising network. When training the parser-free model, we select an element \({\widetilde{P}}_i\) from image set \(S_{\widetilde{P}}\) and use it together with the paired garment image \(G\) as input. To reduce computational complexity for high-resolution image synthesis, we use a frozen pretrained KL-regularized autoencoder~\cite{esser2021taming} to compress person image \({\widetilde{P}}_i\), garment image \(G\) and target \(Z_{P}\) into latent representations \({Z_{\widetilde{P}}},Z_G,Z_{P}\in\mathbb{R}^{C \times H/f \times W/f}\) with down-sampling rate \(f\):

\begin{align}
\left\{
    \begin{aligned}
    &Z_{\widetilde{P}} = \mathcal{E}(\widetilde{P}_i) \\
    &Z_{G} = \mathcal{E}(G) \\
    &Z_{P} = \mathcal{E}(P)
    \end{aligned}
\right.
\end{align}

In the training stage, we perform the diffusion process by adding noise to the latent embedding of the target person
image \(Z_{P}\) according to the noise schedule. The noised latent embedding \(Z_{P}^t\) at diffusion timestep \(t\) could be calculated as below:
\begin{align}
Z_{P}^t=\sqrt{\bar{\alpha_t}} Z_{P}+\sqrt{1-\bar{\alpha_t}} \epsilon
\end{align}
, where \(\bar{\alpha_t}:=\prod_{s=1}^t(1-\beta_s)\) and \(\beta_s\) is the variance in the noise schedule \(\beta\). \(\epsilon\) denotes the Gaussian noise \(\mathcal{N}(0,\textbf{I})\).

Next, \(Z_{P}^t\) and \(Z_{\widetilde{P}}\) are concatenated and feed into the UNet-based~\cite{ronneberger2015u} denoising network, which encodes them and adds the sinusoidal embedding of the timestep  \(t\) as the source of the input block. The garment embedding \(Z_G\) is fed into the garment encoder and then fused into the UNet backbone by the cross-attention mechanism. The noise estimator \(\epsilon_\theta\) is predicted by optimizing the network with a Mean Squared Error (MSE) loss function \(\mathcal{L}\):
\begin{align} 
\mathcal{L}_{\text{MSE}}=E_{t, Z_{P}, \epsilon}\left[\left\|\epsilon-\epsilon_{\theta}\left(Z_{P}^t, t, Z_{\widetilde{P}}, Z_G, \right)\right\|^{2}\right]
\end{align}

\noindent\textbf{Classifier-free Diffusion Guidance.} Early conditional sampling methods, e.g., Guided-Diffusion~\cite{dhariwal2021diffusion}, need to train a classifier to guide the conditional diffusion model. In this paper, we follow the classifier-free diffusion  guidance~\cite{ho2022classifier} strategy, a more elegant way, in which we jointly train an unconditional model and perform the sampling using a linear combination of the conditional and unconditional estimated noises without an external classifier.

In the sampling stage, the model will gradually denoising from the initial noise \(Z_{\hat{P}}^T\sim \mathcal{N}(0,\textbf{I})\). Following the classifier-free guidance, we set \(Z_{\widetilde{P}}, Z_G\) to a zero-filled matrix as unconditional inputs. The final estimated noise \(\hat{\epsilon}_{\theta}\) can be represented as below:
\begin{equation}
\begin{split}
\hat{\epsilon}_{\theta}(Z_{\hat{P}}^t| Z_{\widetilde{P}},& Z_G) =\epsilon_{\theta}(Z_{\hat{P}}^t|\emptyset,\emptyset)\\
&+s\cdot(\epsilon_{\theta}(Z_{\hat{P}}^t| Z_{\widetilde{P}}, Z_G)-\epsilon_{\theta}(Z_{\hat{P}}^t|\emptyset,\emptyset))
\end{split}
\end{equation}
, where \(s\) is the guidance scale and \(t\) is the timestep.

Then, the latent noised person image in the timestep \(t-1\) is calculated as follows
\begin{align} 
Z_{\hat{P}}^{t-1} = \frac{1}{\sqrt{\alpha_t}}(Z_{\hat{P}}^t - \frac{1 - \alpha_t}{\sqrt{1-\beta_t}}\hat{\epsilon}_\theta + \Sigma^{\frac{1}{2}} \mathbf{n})
\end{align}
, where \(\textbf{n} \sim \mathcal{N}(0,\textbf{I})\) and the \(\sigma\) denotes the variance predicted from the model by a Variational Lower Bound (VLB) loss.

After T-step denoising, we obtain the clean latent embedding \(Z_{\hat{P}}^0\) and recover final try-on results \(\hat{P}=\mathcal{D}(Z_{\hat{P}}^0)\) by the pretrained decoder.

\subsection{Garment Feature Extraction and Implicit Warping}
\label{sec:3.2}

\noindent\textbf{Garment Feature Extraction.} To warp and blend the garment to the target person in latent space effectively, their features are either fused by simple concatenating~\cite{morelli2023ladi} or by a two-stream parallel U-Nets~\cite{zhu2023tryondiffusion, gou2023taming}, in which the garment and person stream both have complete U-Net structure and are fused by cross-attention.

Differently, we propose a one-and-a-half stream U-Nets (as shown in Fig.~\ref{fig:architecture}) to do the task. \(Z_G\) is only processed by the encoder (half of U-Net) for feature extraction. Then the multiscale garment features are injected into the person stream U-Net in parallel at the same scales.  In this way, our method can train more efficiently and reduce the number of parameters of the model.

\begin{figure}[tp]
    \centering
    \includegraphics[width=\linewidth]{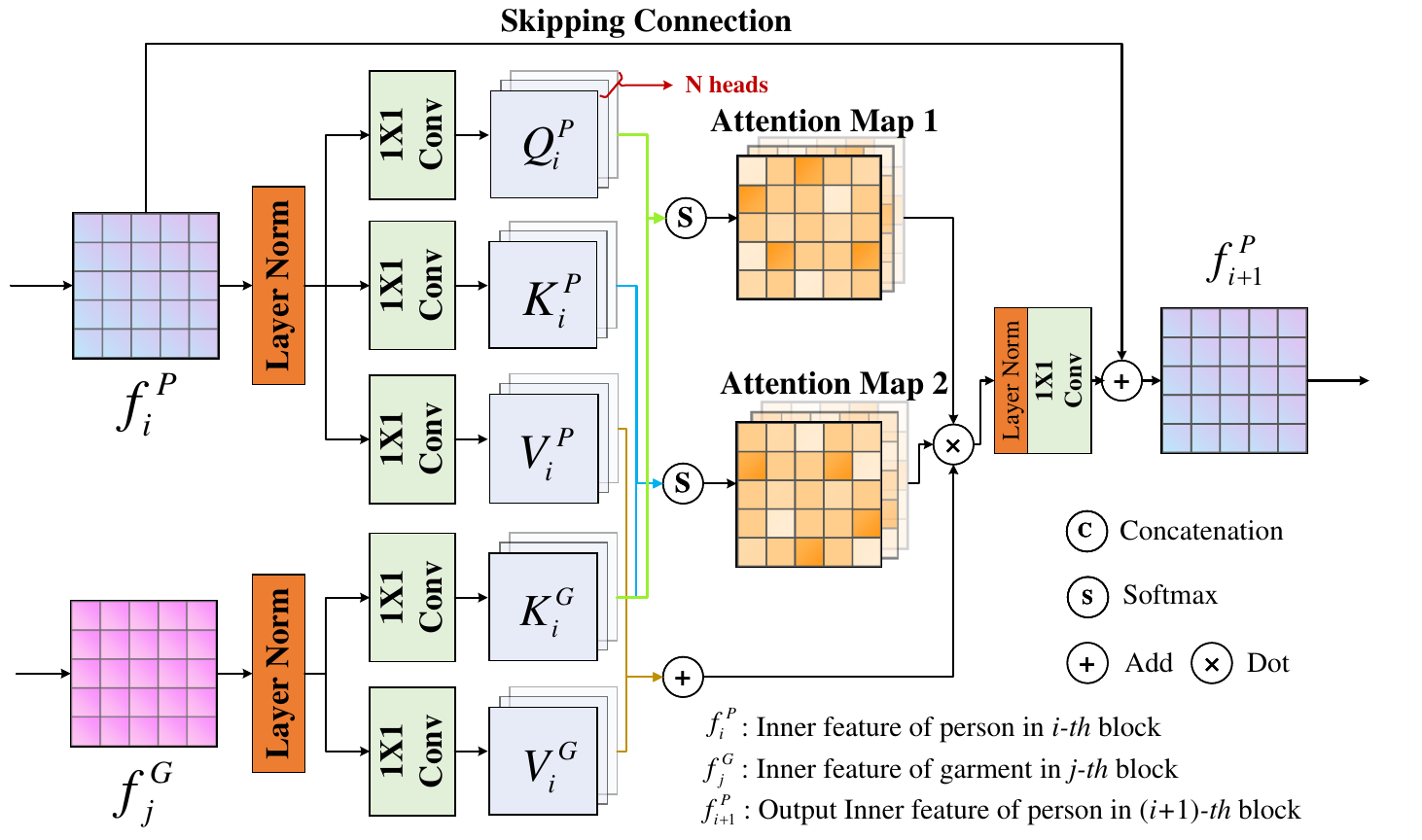}
    \caption{The Garment Fusion Attention Module. Given \(f_i^P, f_j^G\), the enhanced attention is applied to fuse inner features of the person and the garment into output \(f_{i+1}^P\).}
    \label{fig:attention}
\vspace{-.65cm}
\end{figure}

\noindent\textbf{Garment Fusion Attention Module.} To improve the integration of person and garment features, we propose a Garment Fusion Attention (GFA) module, inspired by \cite{chen2023diffusion}. This module is depicted in Fig.~\ref{fig:attention}. Here, \(f_i^P\) and \(f_j^G\) represent the inner features of the person and garment in the \(i\)-th and \(j\)-th blocks, respectively. After layer normalization, we use \(1 \times 1\) convolution to generate query(Q), key(K), and value(V). In detail, \(\{Q_i^P, K_i^P, V_i^P\}\) are generated from \(f_i^P\) and \(\{K_i^G, V_i^G\}\) are generated from \(f_i^G\). To learn more complete representations from various views, we split features into \(N\) heads.  The attention maps are calculated as follows:
\begin{align}
\left\{ 
    \begin{aligned}
    &M_1=Softmax(\frac{Q_i^P(K_i^G)^T}{\sqrt{d}}) \cr 
    &M_2=Softmax(\frac{K_i^P(K_i^G)^T}{\sqrt{d}}) \cr 
    \end{aligned}
\right.
\end{align}
, where \(d\) denotes the number of channel. Then, the final attention function can be described below
\begin{align} 
Att(Q_i^P,K_i^P,V_i^P,K_i^G,V_i^G)=M_1M_2(V_i^P+V_i^G)
\end{align}

\begin{figure*}[tp]
    \centering
    \includegraphics[width=\linewidth]{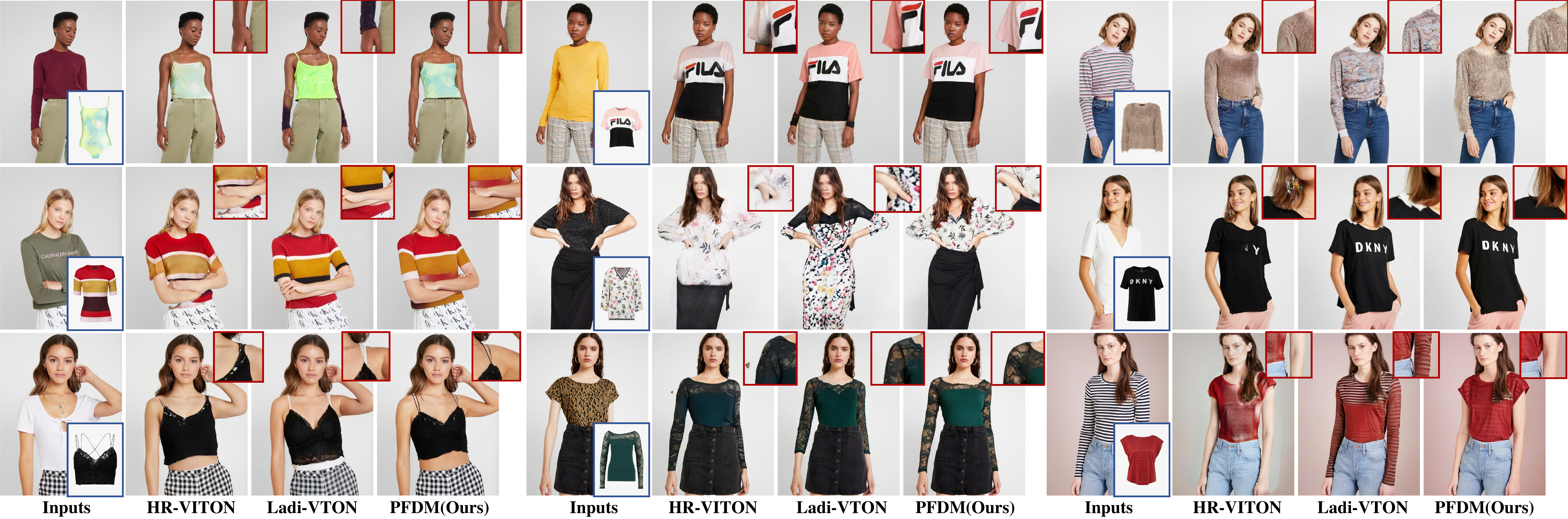}
    \caption{Visual Comparison of PFDM and other competitive methods. From left to right: the reference person and the in-shop garment, images generated by HR-VITON~\cite{lee2022high}, Ladi-VTON~\cite{morelli2023ladi}, PFDM (ours). }
    \label{fig:visual_comparison}
    \vspace{-.4cm}

\end{figure*}
Different from the original cross attention, the values \(V\) which only contains the person image's information, the GFA module includes two groups of attention matrices and values to further enhance the fusion of person and garment features and the expressive ability of the model.

\section{Experiments}

\label{sec:exp}
\subsection{Experimental Setup}
\noindent\textbf{Datasets.} We evaluate our model on VITON-HD dataset~\cite{choi2021viton}, which contains 13,679 high-resolution , i.e., \(1024\times768\), frontal-view woman and upper body clothing image pairs. Following previous works, we divide the dataset into training and test sets with 11,647 and 2,032 pairs. For parser-free model training, we use three well-performed models~\cite{choi2021viton, lee2022high, xie2023gp} to generate pseudo-inputs in \(1024\times768\) resolution. For each model, we generate 10 images of the same person wearing random different garments, thus scaling up the training set 30 times.

\noindent\textbf{Implementation details.} The scale factor \(f\) of pretrained autoencoder is 8 and the spatial dimension of latent space is \(C \times H/f \times W/f\), where channel \(C\) is 4. For U-Net, the channel multiplier for each level of the UNet is set to \((3,4,6,7)\). We make cross-attention connections at the scale of \((32,16)\) with 8 heads. The diffusion model is optimized by Adam optimizer with learning rate \(1e^{-4}\) with 1,000 noise steps. In the inference stage, the model is sampled using DDPM~\cite{ho2020denoising} and the guidance scale \(s\) is 2. All experiments are conducted on NVIDIA Tesla A800s.

\begin{table}[t]
    \centering
    \caption{Quantitative results of unpaired settings in terms of FID and KID on the VITON-HD dataset at 256×192, 512×384 and 1024×768 resolutions. The KID is scaled by 1000 for better comparison.
    }
    \vspace{.12cm}
    \label{tab:fid}
    \setlength{\tabcolsep}{.3em}
    \resizebox{\linewidth}{!}{
    \begin{tabular}{l c cc c cc c cc}
    \toprule

     \multirow{2}{*}{\textbf{Method}} & \multirow{2}{*}{\makecell{\textbf{Parser-free}}} & \multicolumn{2}{c}{\textbf{256×192}} & & \multicolumn{2}{c}{\textbf{512×384}} & & \multicolumn{2}{c}{\textbf{1024×768}} \\
     
    \cmidrule{3-4} \cmidrule{6-7} \cmidrule{9-10}
     
     & & \textbf{FID}\(\downarrow\) & \textbf{KID}\(\downarrow\) & & \textbf{FID}\(\downarrow\) & \textbf{KID}\(\downarrow\) & & \textbf{FID}\(\downarrow\) & \textbf{KID}\(\downarrow\) \\
    
    \midrule
    VITON-HD~\cite{choi2021viton} & N & 16.36 & 8.71 & & 11.64 & 3.00 & & 11.59 & 2.47 \\
    HR-VITON~\cite{lee2022high} & N & 9.38 & 1.53 & & 9.90 & 1.88 & & 10.91 & 1.79 \\
    PF-AFN~\cite{ge2021parser} & Y & 11.49 & 3.19 & & 11.30 & 2.83 & & 14.01  & 5.88 \\
    Ladi-VTON~\cite{morelli2023ladi} & N & 8.23 & 0.96 & & 9.41 & 1.60 & & - & - \\
    DCI-VTON~\cite{gou2023taming} & N & 8.02 & 0.58 & & 8.09 & \textbf{0.28} & & 9.13 & 0.87 \\

    \midrule
    \rowcolor{LightCyan}
    \textbf{PFDM(Ours)} & Y & \textbf{7.38} & \textbf{0.34} & & \textbf{7.99} & 0.38 && \textbf{8.26} & \textbf{0.34}\\

    \bottomrule
    \end{tabular}
    }
\vspace{-.4cm}
\end{table}

\subsection{Experimental Results.} 

\noindent\textbf{Visual Comparison.} We compare our model with images produced by HR-VITON and Ladi-VTON, as shown in Fig.~\ref{fig:visual_comparison}. It's obvious that our model outperforms the state-of-the-art models in terms of clothing details and fitness. Especially for some difficult cases, such as garments in different shapes, persons with complex poses, etc., our models could better preserve the texture details of the clothes, and generate more fitting and natural try-on images.

\noindent\textbf{Quantitative Results.} 
In the paired setting, given reference persons and unpaired garments, we use Frechet Inception Distance (FID) and Kernel Inception Distance (KID) to measure the distributed distance of generated and original images. It can be seen in Tab.~\ref{tab:fid} that PFDM achieves SOTA performance at almost all aspects on VITON-HD. Especially, our model significantly improves KID at \(1024 \times 768\) and \(256 \times 192\) resolutions (from 0.87 to 0.34 and 0.58 to 0.34).

In the unpaired setting, given the pseudo-label of a person wearing another garment and the original paired garment, we generate the original person image. We use Structural Similarity (SSIM) and Learned Perceptual Image Patch Similarity (LPIPS) to measure the similarity of each generated and original image. Compared with the previous SOTA models (Tab.~\ref{tab:other}), our model outperforms them in all aspects and achieves an obvious improvement in KID.

\begin{table}[t]
    \centering
    \caption{Quantitative results of paired settings in terms of FID, KID, LPIPS and SSIM on the VITON-HD dataset at 512×384 resolution. The KID is scaled by 1000.
    }
    \vspace{.12cm}
    \label{tab:other}
    \resizebox{\linewidth}{!}{
    \begin{tabular}{lc c cccc}
    \toprule


    \textbf{Method} & & \textbf{LPIPS} \(\uparrow\) & \textbf{SSIM} \(\uparrow\) & \textbf{FID$_\text{p}$} \(\downarrow\) & \textbf{KID$_\text{p}$} \(\downarrow\) \\

    \midrule
    VITON-HD~\cite{choi2021viton} & & 0.116 & 0.863 & 11.01 & 3.71  \\
    HR-VITON~\cite{lee2022high} & & 0.097 & 0.878 & 10.88 & 4.48 \\
    Ladi-VTON~\cite{morelli2023ladi} & & 0.091 & 0.876 & 6.66 & 1.08 \\

    \midrule
    \rowcolor{LightCyan}
    \textbf{PFDM(Ours)} & & \textbf{0.076} & \textbf{0.891} & \textbf{4.99} & \textbf{0.19} \\

    \bottomrule
    \end{tabular}}
    
\vspace{-.55cm}
\end{table}

\noindent\textbf{Ablation study.}
To evaluate the effectiveness of the key steps in our model, we conduct the ablation study on the VITON-HD dataset at the original resolution in the paired setting, and the results are shown in Tab.~\ref{tab:ablation}. The baseline is a primitive model with the vanilla cross-attention module training only on one-model-generated pseudo-inputs, and no guidance technique is used for inference. Subsequently, we investigated how the results are influenced by the three conditions. We added multi-model pseudo-label generation(MP), classifier-free guidance technique(CF), and garment fusion attention(GFA) modules to the baseline one by one. The experiments show that applying these modules or methods results in obvious performance improvement.

\begin{table}[h]
    \vspace{-.22cm}
    \fontsize{9pt}{9pt}\selectfont
    \centering
    \caption{Quantitative comparison for ablation studies. We compute FID and KID on VITON-HD at 1024×768 resolution. The KID is scaled by 1000. }
    \vspace{.12cm}
    \label{tab:ablation}
    \linespread{1.5}
    \resizebox{0.8\linewidth}{!}{
    \begin{tabular}{lc cc}
    \toprule
     
    \textbf{Model} & & \textbf{FID} $\downarrow$ & \textbf{KID} $\downarrow$ \\
    \midrule
     baseline& & 9.63 & 1.12 \\
     baseline+MP& & 8.65 & 0.54\\
     baseline+MP+CF& & 8.36 & 0.37 \\
    \midrule
    \rowcolor{LightCyan}
    \textbf{baseline+MP+CF+GFA(Ours)} & & \textbf{8.26} & \textbf{0.34} \\

    \bottomrule
    \end{tabular}
    }
\vspace{-.25cm}

\end{table}

\section{Conclusion}
\label{sec:conclusion}
In this paper, we proposed a parser-free virtual try-on method based on diffusion model, which unified the warping and blending steps into one model while avoiding the use of any parser or external module. As we know, PFDM is the first diffusion-based parser-free model for virtual try-on. Experiments show that PFDM can generate high-fidelity try-on results in high resolution with rich texture details and successfully handle misalignment and occlusion, which not only outperforms the existing parser-free methods but also surpasses state-of-the-art parser-based models in both qualitative and quantitative analysis. We hope that our work could promote the popularization of virtual try-on technology in e-commerce and meta-verse.

\vfill\pagebreak


\bibliographystyle{IEEEbib}
\bibliography{refs}

\end{document}